\documentclass{article}

\usepackage{microtype}
\usepackage{graphicx}
\usepackage{subcaption}
\usepackage{amsmath,amssymb,amsfonts}
\usepackage{booktabs} 
\usepackage{xcolor, enumitem}

\usepackage{hyperref}

\setlist{nosep}

\usepackage[accepted]{icml2020}

\icmltitlerunning{The Battlesnake challenge}

\begin{document}

\twocolumn[
\icmltitle{Battlesnake Challenge: A Multi-agent Reinforcement Learning Playground with Human-in-the-loop}



\icmlsetsymbol{equal}{*}

\begin{icmlauthorlist}
\icmlauthor{Jonathan Chung}{equal,am}
\icmlauthor{Anna Luo}{equal,am}
\icmlauthor{Xavier Raffin}{am}
\icmlauthor{Scott Perry}{am}
\end{icmlauthorlist}

\icmlaffiliation{am}{Amazon Web Services}

\icmlcorrespondingauthor{jonchung, annaluo}{@amazon.com}

\icmlkeywords{Reinforcement learning, Human-in-the-loop learning}

\vskip 0.3in
]



\printAffiliationsAndNotice{\icmlEqualContribution} 

\begin{abstract}
We present the Battlesnake Challenge, a framework for multi-agent reinforcement learning with Human-In-the-Loop Learning (HILL).
It is developed upon \textit{Battlesnake}, a multiplayer extension of the traditional \textit{Snake} game in which 2 or more snakes compete for the final survival.
The Battlesnake Challenge consists of an offline module for model training and an online module for live competitions.
We develop a simulated game environment for the offline multi-agent model training and identify a set of baseline heuristics that can be instilled to improve learning.
Our framework is \textit{agent-agnostic} and \textit{heuristics-agnostic} such that researchers can design their own algorithms, train their models, and demonstrate in the online Battlesnake competition. 
We validate the framework and baseline heuristics with our preliminary experiments.
Our results show that agents with the proposed HILL methods consistently outperform agents without HILL. Besides, heuristics of reward manipulation had the best performance in the online competition.
We open source our framework at https://github.com/awslabs/sagemaker-battlesnake-ai.
\end{abstract}

\section{Introduction}

Battlesnake is an extension of the traditional \textit{Snake} arcade game where multiple snakes compete against one another for food and survival.
The last surviving snake is the winner of the game. 
Competitors traditionally develop heuristics such as using the A* search algorithm \cite{russell2002artificial} and the tree search \cite{mci/Schier2019} to seek food, enemy heads, and its tail.
Meanwhile, Reinforcement Learning (RL), which learns a policy by interacting with an environment through trial-and-error, has been naturally adopted to tackle such sequential problems. 
Recent advances in deep RL further allows modelling such decision making problems with high-dimensional visual perceptual inputs made up of thousands of pixels \cite{mnih2015human}.
Battlesnake focuses on a particular branch of RL where multiple agents learn to interact within the same environment.
It is denoted as multi-agent RL problems  \cite{littman1994markov, bu2008comprehensive, bucsoniu2010multi}. 
The game fits the fully competitive setting \cite{silver2017mastering} in which each agent is tasked to maximise its own reward while minimising their opponent's rewards.

Developers with superior domain knowledge build snakes with unique strategies and heuristics.
Having Human-In-the-Loop Learning (HILL) aids the policy training and prevents catastrophic actions \cite{christiano2017deep, abel2017agent}.
Human intuition could be provided as feedback \cite{arakawa2018dqn,xiao2020fresh}, teachers \cite{abel2017agent,zhang2019leveraging}, and overseers \cite{saunders2018trial}.
It can also steer agents to have more optimal learning by identifying important and omitting misleading features, subsequently reducing the dimensionality of the state space \cite{abel2017agent}.
However, these methods incorporate human guidance in a single agent setting.
To the best of our knowledge, there exists no playground designated to evaluate multi-agent RL algorithms with HILL.

To fill in this gap, we introduce the Battlesnake Challenge, an accessible and standardised framework that allows researchers to effectively train and evaluate their multi-agent RL algorithms with various HILL methods.
We choose to use Battlesnake as the underlying game engine because it can be seamlessly integrated to the multi-agent RL research direction, while remaining intuitive and lightweight. Besides, the rules governing the game are relatively simple, but the  resulting strategies can still be complex. Such setting facilitates the use of human knowledge to provide policy training guidance.
In specific, we offer a simulated Battlesnake environment for offline training, after which the snakes can be deployed to the cloud to compete with other snakes in the Battlesnake global arena\footnote{https://play.battlesnake.com/arena/global/. The arena hosts all types of AI bots, being RL or non-RL, to compete against each other.}.
To accommodate human knowledge, we identify a number of standard heuristics and further demonstrate the baseline techniques to incorporate them into the agent.
Our proposed framework is \textit{agent-agnostic} and \textit{heuristics-agnostic} such that researchers can design their own algorithms, train their models, and demonstrate in the real Battlesnake competition.
We hope that the Battlesnake Challenge will serve as a testbed to encourage research into multi-agent RL and HILL.
Our contributions are as follows:
\begin{itemize}
    \item We propose an end-to-end training-deployment-testing framework that consists of an offline module for multi-agent RL training with HILL, and an online module for performance evaluation against other public agents.
    \item On the multi-agent RL aspect, we develop a simulator for the Battlesnake problem through a proper design of state, action and reward. On HILL aspect, we identify a set of baseline heuristics that can be instilled to improve the agents during training or after deployment.
    \item We validate the proposed framework and baseline heuristics with our preliminary experiments.
    We investigate how different incorporation methods affect task performance both offline and online. 
    Our results show that agents with HILL outperform agents without HILL and careful reward manipulation performs the best among our proposed heuristics in the online Battlesnake arena.
    \item We open-source the Battlesnake Challenge framework at \texttt{http://github.com/ awslabs/sagemaker-battlesnake-ai} to encourage broader research directions.
\end{itemize}

\section{Related works}
\textbf{Multi-agent RL Testbed:} There is a growing number of studies focusing on designing environments to evaluate the agents' performance with the advancements in the multi-agent RL regime \cite{zhang2019multi, nguyen2020deep}.
For example, Keepaway soccer \cite{stone2005keepaway} and its extension \cite{kalyanakrishnan2006half,hausknecht2016half} provide a simulated football environment.
Meanwhile, a set of gridwold-like environments has been developed to encompass various multi-agent tasks, covering both continuous \cite{lowe2017multi} and discrete \cite{yang2018mean, zheng2018magent} control problems. 
\citet{resnick2018pommerman} proposed \textit{Pommerman}, 
a game stylistically similar to the Nintendo game Bomberman, 
as a playground for bench-marking different agents.
The system uses low dimensional symbolic state interpretations input, and the authors built an online leader board where researchers could submit their agents and compete against one another. 
It, however, allows only up to four agents and does not include the mechanism that adds human intuition to the agents.

More recent work considers real-time strategy games that require complex environments and controls. 
The StarCraft Multi-Agent Challenge (SMAC) \cite{samvelyan2019starcraft, vinyals2019grandmaster}, developed based on \textit{StarCraft II}, focuses on handling partially observability and high-dimensional inputs. 
It aims to serve as a benchmark for \textit{cooperative} multi-agent RL, rather than the \textit{competitive} setting as in Battlesnake. 
In contrast, \citet{ye2019mastering} presented a 1v1 game mode using multi-agent RL in a competitive setting through the game \textit{Honor of Kings}. 
They developed an off-policy RL system architecture for scalable training and eliminated invalid actions to improve the training efficiency.
Nonetheless, the complexities inherent in both of these games require specific game knowledge, making it difficult for general researchers to develop and evaluate their multi-agent RL algorithms.

\textbf{Human-in-the-loop Reinforcement Learning:} The data hungry nature of RL has prompted researchers to develop techniques to leverage human knowledge for RL tasks \cite{zhang2019leveraging}. 
Often, human information is passed along in the form of human intervention \cite{saunders2018trial}, reward shaping \cite{knox2008tamer, knox2009interactively,knox2012reinforcement,warnell2018deep,arakawa2018dqn,xiao2020fresh} and policy evaluation \cite{griffith2013policy,macglashan2017interactive,arumugam2019deep}.
In principal, human intuition could be injected in two methods, 1) by evaluating the actions during training through real-time feedback or intervention; and 2) by defining handcrafted rules prior to training to alter the agent's behaviour based on human intuition.
TAMER+RL \cite{knox2010combining} is an example of the first method in which human provide a reward given a state action pair.
An example of the second method is the agent-agnostic framework proposed by \citet{abel2017agent}.
The framework contains a protocol program that controls the human interaction with a single agent.
Human can perform learning interventions with handcrafted rules that alter the transition dynamics given the current state and action, with methods including action masking, reward manipulation, etc.
Action masking is a technique to bar possible actions from the policy based on engineered rules.
Reward manipulation is the act of specifically designing a reward function.
In particular, the authors found that action masking simplifies the RL problem and was effective to prevent catastrophic actions \cite{saunders2018trial}.
Considering that Battlesnakes are traditionally developed with handcrafted rules, we extend the framework into a multi-agent RL setting.

\section{Description of the Battlesnake Challenge}

\begin{figure}[h!]
\centering
\includegraphics[width=0.43\textwidth]{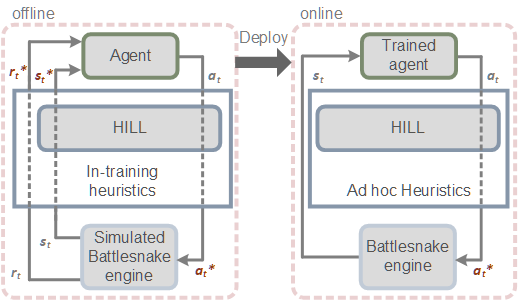}
\caption{Overview of the Battlesnake Challenge with human-in-the-loop multi-agent reinforcement learning. Human knowledge injected in the offline training phase can possibly affect state ($s_t$), action ($a_t$) and reward ($r_t$) at timestep $t$. Once the agents is deployed, Ad-hoc heuristics can overwrite the action ($a_t$) at each inference timestep $t$.}
\label{fig:battlesnake_framework}
\end{figure}

The architecture design of the Battlesnake Challenge is presented in Figure \ref{fig:battlesnake_framework}.
Our framework includes an offline RL training module equipped with a simulated Battlesnake environment,
as will be described in Section \ref{subsec:env}.
The framework is designed to work with handcrafted rules \cite{abel2017agent} as well as more complex heuristics \cite{russell2002artificial,mci/Schier2019} that are programmed in advance.
The preprogrammed heuristics are independent of the RL algorithm and could be incorporated into the RL training module to assist the training procedures.

Once the trained agent is deployed online, it can make inferences and interface with the Battlesnake engine to play in the Battlesnake arena.
We integrate the inference with optional ad-hoc heuristics to enable action overwriting, through which human experts can perform the safety checks.
We use the Battlesnake arena to evaluate different combinations of in-training and ad-hoc human guidance by allowing the agents to compete against each other.

\subsection{Battlesnake description}
\label{subsec:env}
We first provide a detailed description of Battlesnake game logic.
A typical game in Battlesnake consists of three to five snakes on a board ranging from $7 \times 7$ (small), $11 \times 11$ (medium) to $19 \times 19$ (large).
At the start of the game, $N$ snakes are randomly distributed along the boundaries of the board, each with health of $100$. 
There is one piece of food randomly distributed at the same time.
At each turn, the health of every snake decreases by one and each snake reacts to the environment indicating whether it will move up, down, left or right;
food are then randomly spawned.
Unlike the traditional snakes game, if a snake is facing up and the next action is to move down, the game considers the snake hitting its own body and it will be eliminated from the game.
This is known as a \textit{forbidden move}.
If a snake eats a food, its health will be returned to 100 and its length will grow by one.
If a snake hits another snake's head, the shorter of the two snakes is eliminated from the game (if the sizes of the two snakes are the same, both the snakes are eliminated).
This is referred to as \textit{eating another snake}.
In addition, a snake is eliminated from the game if it: 1) goes out of the boundaries of the map, 2) hits another snake's body, 3) hits its own body, or 4) has a health of 0.
The last surviving snake becomes the winner.

\subsection{Battlesnake as a Reinforcement Learning Environment}
\label{sec:bs_rl}
\begin{figure}[h!]
\centering
\includegraphics[width=0.4\textwidth]{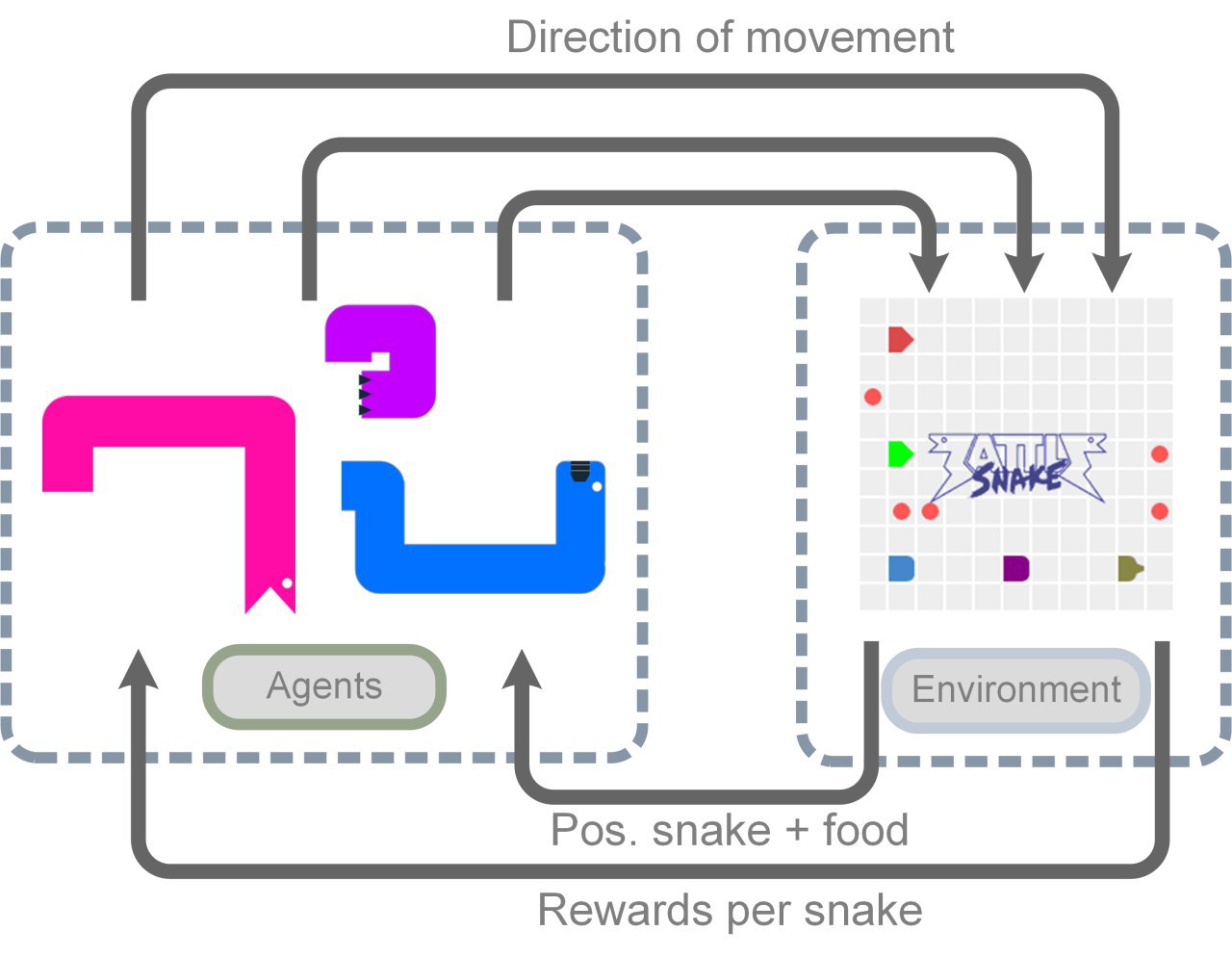}
\caption{Modelling Battlesnake with reinforcement learning. 
}
\label{fig:rl_battlesnake}
\end{figure}

We consider a standard Markov game \cite{littman1994markov} to model the interaction between multiple Battlesnake agents with the environment.
Each agent represents one snake in the game.
The Markov game is specified by a tuple $M = (\mathcal{N}, \mathcal{S},\{\mathcal{A}^i\}_{i\in\mathcal{N}},\mathcal{T},\{R^i\}_{i\in\mathcal{N}},\gamma)$, where $\mathcal{N} = \{1,\dots,N\}, N> 1$ denotes the set of agents, 
$\mathcal{S}$ is the state space observed by all agents and $\mathcal{A}^i$ is the action space of agent $i$. 
$\mathcal{T}: \mathcal{S}\times\mathcal{A}^1\times\cdots\mathcal{A}^N\times\mathcal{S} \rightarrow [0,1]$ denotes the transition function that maps a state $s_t\in \mathcal{S}$ and action $a_t^i\in\mathcal{A}^i$ pair for each agent $i$ to a probability distribution over the next state $s_{t+1} \in \mathcal{S}$. 
The environment emits a reward $R^i: \mathcal{S}\times\mathcal{A}^1\times\cdots\mathcal{A}^N\rightarrow \mathbb{R}$ on each transition for each agent $i$; $\gamma$ denotes the discount factor.
Figure \ref{fig:rl_battlesnake} illustrates setup in which the agent interacts with the simulator over the OpenAI Gym interface \cite{brockman2016openai}. Components in the Markov game are given as follows:

\textbf{State:} We provide the Battlesnake simulator a gridworld based state space $s_t$ at time $t$ to represent the spatial distribution of all the snakes and food.
Agent $i$ is represented by a list of coordinates, $\textbf{x}^i = [x_1^i, x_2^i, \dots, x_{L_i}^i]$ where $x_j^i \in \mathbb{R}^2$ $\forall j \in 1, \dots, L_i$ and $L_i$ is the length of snake $i$.
The $N$ snakes are collectively referred to with $\textbf{X}$ such that $\textbf{X} = [\textbf{x}^1 ... \textbf{x}^i ...\textbf{x}^N]$.
Food $\textbf{F}$ is represented as $[y^1, y^2 \dots y^{M^t}]$ where $y^j\in\mathbb{R}^2$ is the coordinate corresponding to the location of the food $j$ and $M^t$ represents the number of food at time $t$.
To ingest the information of other agents, we organise the state for agent $i$, $s_{t}^{i}$ as a grid with 3 channels where $s_{t}^{i} \in \mathbb{R}^{w \times h \times 3}$, and $w$ and $h$ are the width and height of the map.
Specifically, assume $s_{t}^{i}[j, k, c]$ describes the state value at coordinate $(j, k)$ on the map of channel $c$ for agent $i$ at time $t$.
Channels $c \in \{0, 1, 2\}$ represents position of food, agent $i$, and other agents, respectively.
For $c = 0$, $s_{t}^{i}[j, k, 0] = 1$ if $(j, k) \in \textbf{F}$ and $0$ otherwise. 
The state for $c = 0$ should be identical for all agents.
Similarly, $c = 1$ provides the position of agent $i$ where $s_{t}^{i}[j, k, 1] = 1, \forall (j, k) \in \textbf{x}^i$ and $0$ otherwise.
Also, we set $s_{t}^{i}[j_h, k_h, 1] = 5$ where ($j_h, k_h$) denotes the head of agent $i$.
We choose $5$ experimentally to provide a larger differentiation between the body and the head of the snake.
Finally, $c = 2$ is defined as $s_{t}^{i}[j, k, 2] = 1, \forall (j, k) \in \textbf{x}^{i'}$ where $\textbf{x}^{i'}$ are all other agents in $\textbf{X}$ where $i' \ne i$, and the heads of the snakes in $\textbf{x}^{i'}$ are set to $5$ as well.
It is worth mentioning that we deliberately choose to formulate the state space in a gridworld fashion rather than using image pixels or complex embedded features. We believe this will make the RL training easier and encourage research focus on developing the heuristics with HILL.

\textbf{Action:} The action space $\mathcal{A}^i$ is identical for each agent $i$, representing the direction the agent moves towards in the next turn.
Namely, $a_t^i = [0, 1, 2, 3]$ corresponds to up, down, left, and right at time $t$ for agent $i$.
Thus the joint action space is defined as $\textbf{a}_t = [a_t^1, a_t^2 ... a_t^N]$.

\textbf{Reward:} The overall goal is to become the last surviving snake. 
We by default apply the same reward formulation for all the agents. 
Specifically, a negative reward $-1$ is imposed if a snake dies,
and the last snake alive is assigned with a reward $1$.
We grant a small reward $\epsilon=0.002$ for each snake when it survives another turn, with an intuition to promote the snakes to move and grow. 

\subsection{Training Algorithm}
We train each snake's policy independently using the Proximal Policy Optimisation (PPO) algorithm \cite{schulman2017proximal}. It is a widely used, modern on-policy actor-critic algorithm that has presented stable performances in many of the recent successes in deep RL. The algorithm employs two neural networks during training -- a policy network and a value network.
The policy network interacts with the Battlesnake environment and generates Gaussian-distributed actions given the state. The value network estimates the expected cumulative discounted reward using the generalised advantage algorithm \cite{schulman2015high}.
Note that while we use PPO to conduct experiments, the proposed framework is \textit{agent-agnostic} and can fit various discrete action-based state-of-the-art algorithms such as QMIX \cite{rashid2018qmix} and SAC \cite{haarnoja2018soft}.

\subsection{Heuristics with human-in-the-loop learning}
\label{sec:heuristics}

In principal, our proposed framework is \textit{heuristic-agnostic} such that researchers can tackle various heuristic development and bring them into the agent training process.
For the purpose of illustration, in this work we identify several human engineered heuristics that we used as part of the challenge.
The philosophy is to provide agents information regarding what actions to \textit{avoid} and to guide agents towards superficial skills such as heading to food when starving.
Specifically, we provide the following heuristics:
\begin{enumerate}[topsep=0pt,itemsep=-1ex,partopsep=1ex,parsep=1ex]
    \item Avoid hitting the walls.
    \item Avoiding moving in the opposite direction as the snake is facing (\textit{forbidden moves}).
    \item Moving towards and eating food when the snake health is low to prevent \textit{starving}.
    \item Killing another snake (e.g., eating another snake or trapping another snake).
\end{enumerate}

\begin{table}[h!]
\centering
\begin{tabular}{c | c c c} 
 \hline
 Rule & Prevention/ & Interaction &  Training \\ 
      &Promotion  &             &  phase \\ [0.5ex] 
 \hline\hline
 1 & Prev. & Env. & Early \\ 
 2 & Prev. & Env. & Early \\
 3 & Promo. & Env. & Middle \\
 4 & Promo. & Agents & Late \\ [1ex] 
 \hline
\end{tabular}
\caption{Properties of the heuristics. Prev. refers to action prevention; Promo. refers to action promotion. Interaction describes whether the rules are interacting with the environment or other agents. Training phase indicates when the rules become more important.}
\label{table:rule_properties}
\end{table}

Table \ref{table:rule_properties} provides an overview of the properties of each rule.
Prevention/promotion refers to \textit{action prevention} or \textit{action promotion}.
Action prevention rules help eliminate unreasonable actions, similar to the use of catastrophic actions prevention \cite{abel2017agent,saunders2018trial} and action filters \cite{gao2019skynet,meisheri2019accelerating}. 
In most cases, action prevention rules could be resolved with a single conditional statement.
Meanwhile, the described action promotion rules are more complex as they require multiple steps to achieve.
Interaction describes whether the rules are interacting with the environment or other agents.
For example, to manage rule 4, an agent would have to anticipate the movements of other agents in order to kill them.
Finally, training phase indicates when the rules become more important.
Rules 1 and 2 are necessary for basic navigation and movement; 
they prevent the agents from committing ``suicide" in the early phases of training and are essential throughout the duration of training.
Rules 3 is necessary for survival after the early phases of training when basic navigation is learnt.
Rule 4 requires high level strategies once the snakes have no issues with surviving.

While human rules are instilled with a goal to accelerate the learning procedure, they can also be biased and limiting the snakes' performance.
For instance, rule 3 could lead to snakes focusing too much on food, whereas rule 4 could result in over-aggressive snakes.
To this end, we design our platform such that the impact of the heuristics can be controlled and even removed once an agent acquires some basic skills.
Such heuristic impact break-down can also be phrased as a curriculum learning method where a logical ordering or hierarchy of simple skills is learnt during training \cite{matiisen2019teacher, portelas2019teacher}.
We now describe three methods to include the heuristics rules into the RL learning.

\begin{figure}[h]
    \centering
         \centering
         \includegraphics[width=0.4\textwidth]{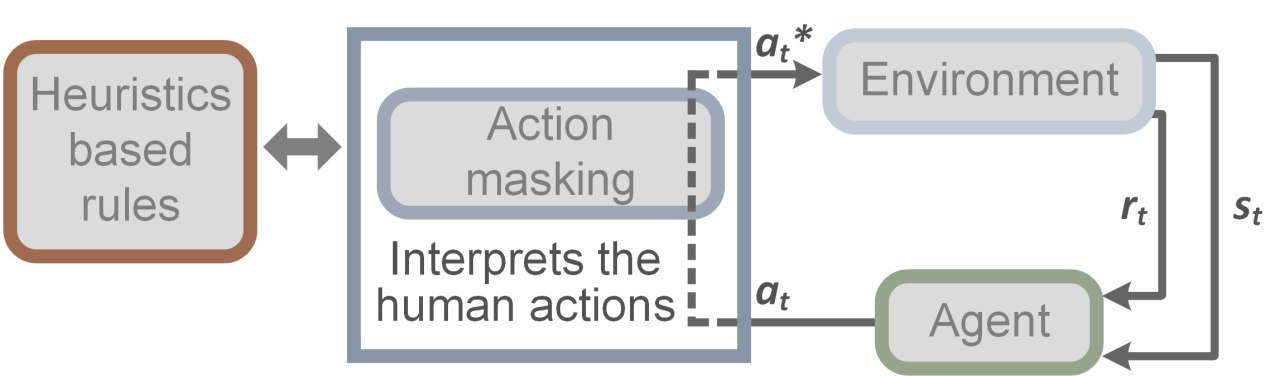}
    \caption{In-training action masking. }
    \label{fig:action_masking}
\end{figure}
\textbf{In-training action masking: }
We first consider the case where the human heuristics are injected during training and the agent adjusts its policy accordingly (Figure \ref{fig:action_masking}).
In particular, we incorporate the feedback at the final output layers of the policy, where invalid actions are masked out of the softmax by scaling the probability to zero.
The actual action $a_t^*$ taken after the masking is then passed into the simulated environment to generate the new states.
Action masking applies to catastrophic action prevention and single step heuristics.
As such we apply it for rules 1 and 2.


\begin{figure}[h]
    \centering
         \centering
         \includegraphics[width=0.4\textwidth]{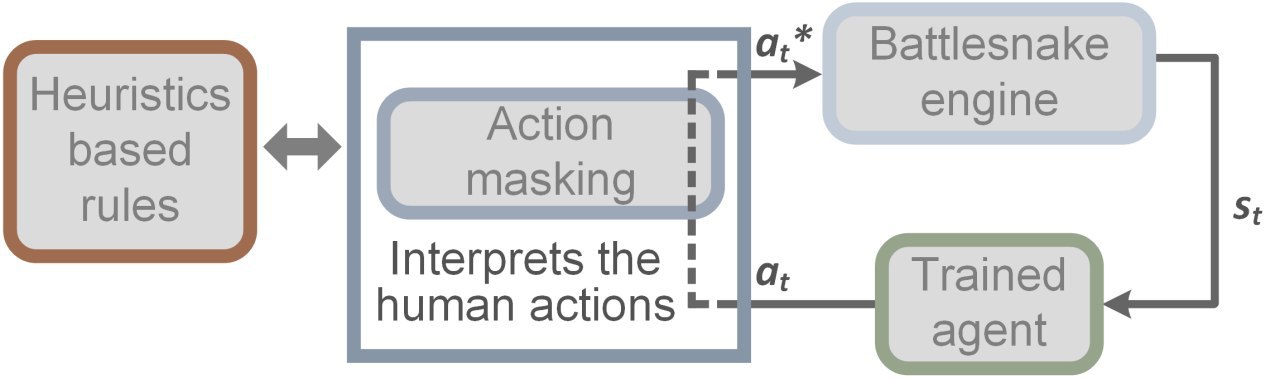}
    \caption{Ad-hoc action overwriting}
    \label{fig:ad_hoc_action_masking}
\end{figure}
\textbf{Ad-hoc action overwriting: }
A trained agent can be deployed to interact with the Battlesnake Engine and compete with other snakes.
Ad-hoc action overwriting is often applied in this scenario to enhance robustness and guarantee performance. 
As shown in Figure \ref{fig:ad_hoc_action_masking}, it is different from in-training action masking in the sense that the heuristics are only applied during inference. 
In our proposed Battlesnake Challenge framework, the actual actions taken and corresponding next states are not used to update the policy in real-time as the agent is already deployed. However, the experiences can be stored for the purpose of policy evaluation.

\textbf{Reward manipulation:}
Reward manipulation includes human intuition by specifically designing a reward function to encourage events corresponding to the heuristics.
During training, the reward function is defined as $\hat{R}(s_t^i, a_t^i) = \hat{r}_t^i$ where $\hat{r}_t^i$ denotes the heuristics based reward.
$\hat{r}_t^i$ is then fed into the learning process with the final reward function defined as $r_t^{*i} = r_t^i + \hat{r}_t^i$.
For instance, we add a penalty term ($\hat{r}_t^i = -0.4$) in our experiments whenever a snake hits a wall to account for rule 1.
Please note that the sign of $\hat{r}_t^i$ changes if the heuristics is action promoting.

\section{Experiments}
\subsection{Implementation details}
We open-source our proposed framework and provide implementation for each component presented in Figure \ref{fig:battlesnake_framework}.
Specifically, our package is featured with a simulated gym environment, a heuristics developer, and the orchestration code to deploy trained agents to the Battlesnake arena.
In addition, we provide a suite of training examples using the RL package RLlib \cite{liang2017rllib} within the Amazon SageMaker RL package.

\subsection{Evaluation}
There are two main avenues to evaluate the RL agent with HILL.
The first avenue evaluates agent performance during training.
The second avenue is to use the leader board in the Battlesnake arena, in which a deployed agent competes against other snakes with black-box mechanisms.

For the first avenue, we collect the maximum \textit{episode length} of all agents as a metric for evaluation.
Episode length provides an indication of how long the agents survived for, and it's consistent across different reward manipulation schemes.
We first investigate the baseline performance with map sizes of $7 \times 7$, $11 \times 11$, and $19 \times 19$ comparing the performances of training with 4, 5, and 6 agents without human intuition.
We aim to verify that the our simulator is properly formulated such that the snake's performance improves over training.
We then compare the performances of heuristic incorporation methods described in Section \ref{sec:heuristics}.
In addition to the episode length, we also record the frequency of events that each heuristics is designed to prevent or encourage to evaluate how well the heuristics are incorporated. 
Here and on-wards, we fix the map size and number of agents at $11 \times 11$ and 5 respectively.

For the second avenue, we evaluate the performances of the described baseline heuristics in the arena. We wish to call out that we do not compare with the existing mature snakes. 
The focus of this study is to showcase how each component in our proposed framework can be modified with a minimum amount of efforts, rather than to develop a sophisticated fine-tuned agent that beats the best performing snakes.
Particularly, we use rule 2 as an example to compare the performance of the trained agents with in-training action masking, ad-hoc action overwriting, reward manipulation, and vanilla training with no HILL.
For each baseline heuristics, we randomly select one policy for deployment.
Note that the ad-hoc action overwriting agent uses the vanilla trained agent as the base agent.
For the in-training action masking agent, we apply the same masking logic during inference to ensure consistency in the state transition dynamics.
We conduct two experiments for the arena testing.
The first experiment consists of $30$ games with the four snakes in the arena.
The last surviving snake is given four points, the second last surviving snake is given three points, and so on.
We also record the frequency of forbidden moves.
In the second experiment, the four snakes compete in a 1 vs. 1 format.
Each pair of snakes plays $10$ games, leading to a total of 60 games.
The winning snake gets 1 point and the losing snake gets 0 point.

\section{Results}
In this section, we present experimental results for components included in the Battlesnke Challenge.
We first demonstrate the performance of the multi-agent RL agents trained in our simulated environment, and then move to performance evaluation with HILL both offline during training and online in the arena.

\begin{figure}[h!]
    \centering
    \includegraphics[width=0.32\textwidth]{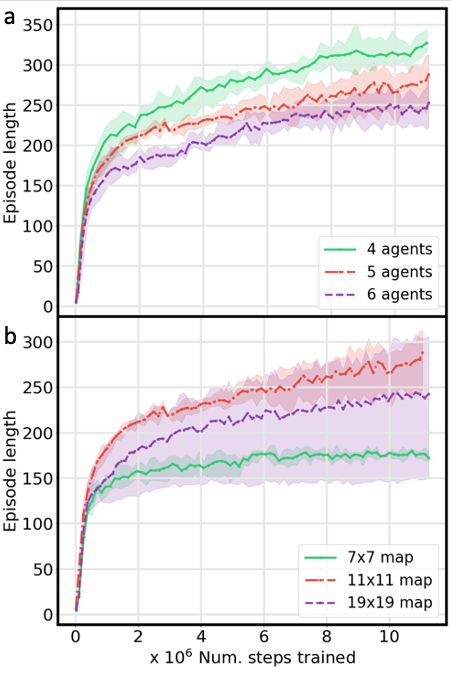}
    \caption{Episode lengths with (a) varying the number of agents on a $11 \times 11$ map and (b) varying the map size with five agents.}
    \label{fig:results_MARL}
\end{figure}

\textbf{Multi-agent reinforcement learning:}
The results of varying the number of agents and the map size are presented in Figure \ref{fig:results_MARL}.
We train three different instances of each case with different random seeds.
The solid curves correspond to the mean and the shaded region to the minimum and maximum values over the trials.
We observe that after 1M steps of training, the episode length increases from 0 to an average of 175. 
After training for 2M steps, the growth increases steadily.
We can observe that the maximum episode length for 4 agents is distinctively better than 5 and 6 agents.
This is not surprising as games with 4 agents have more space to roam around the map compared to games with 5 or 6 agents.
Similarly, when investigating the effects of the map sizes, we observe that the episode length of 5 agents on larger map sizes is longer than that of smaller map sizes.

\subsection{Human-in-the-loop during training}
\begin{figure}[h!]
    \centering
    \includegraphics[width=0.35\textwidth]{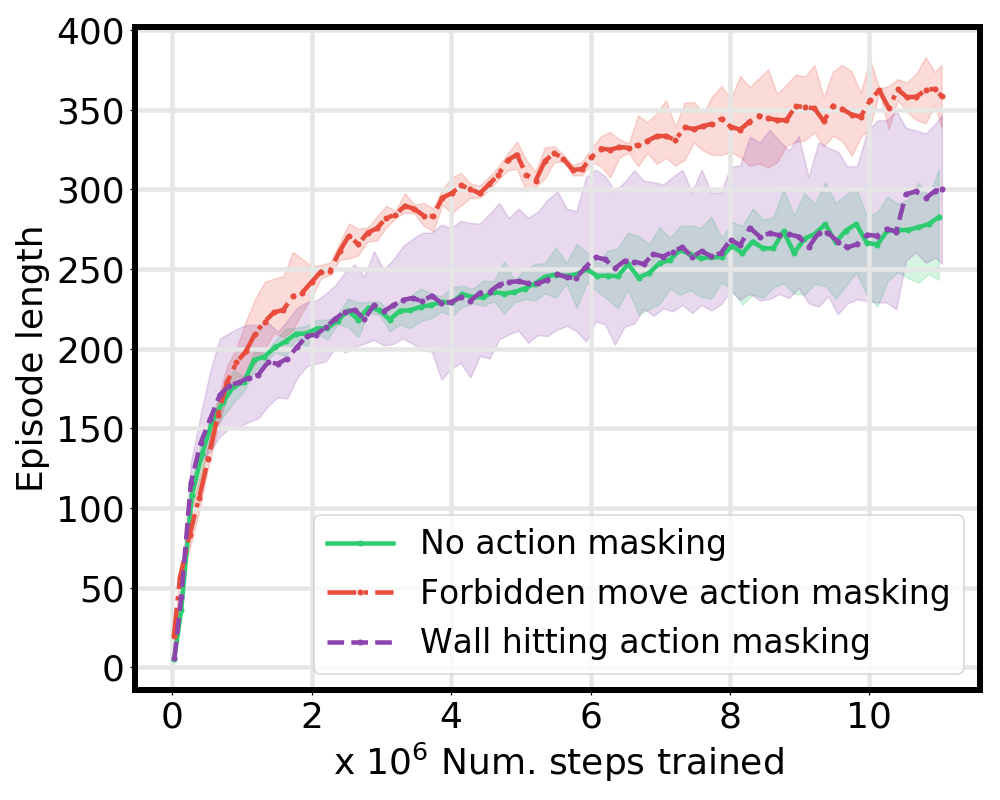}
    \caption{Experiments with action masking for rule 1 and 2 with 5 agents on a $11 \times 11$ map}
    \label{fig:results_AM}
\end{figure}

\textbf{In-training action masking:}
Our results as presented in Figure \ref{fig:results_AM} show that the agents with in-training action masking outperforms the one without it.
Action masking with rule 2 (forbidden moves) has the best performance, reaching an episode length of more than 350 at 10M steps of training.
Action masking with rule 1 (wall hitting) achieves an episode length of about 300 at 10M steps whereas the agent with no heuristics achieves a bit more than 250.
This verifies that including action masking improves the sample efficiency and thus accelerate the policy training.

\begin{figure}[h!]
    \centering
    \includegraphics[width=0.35\textwidth]{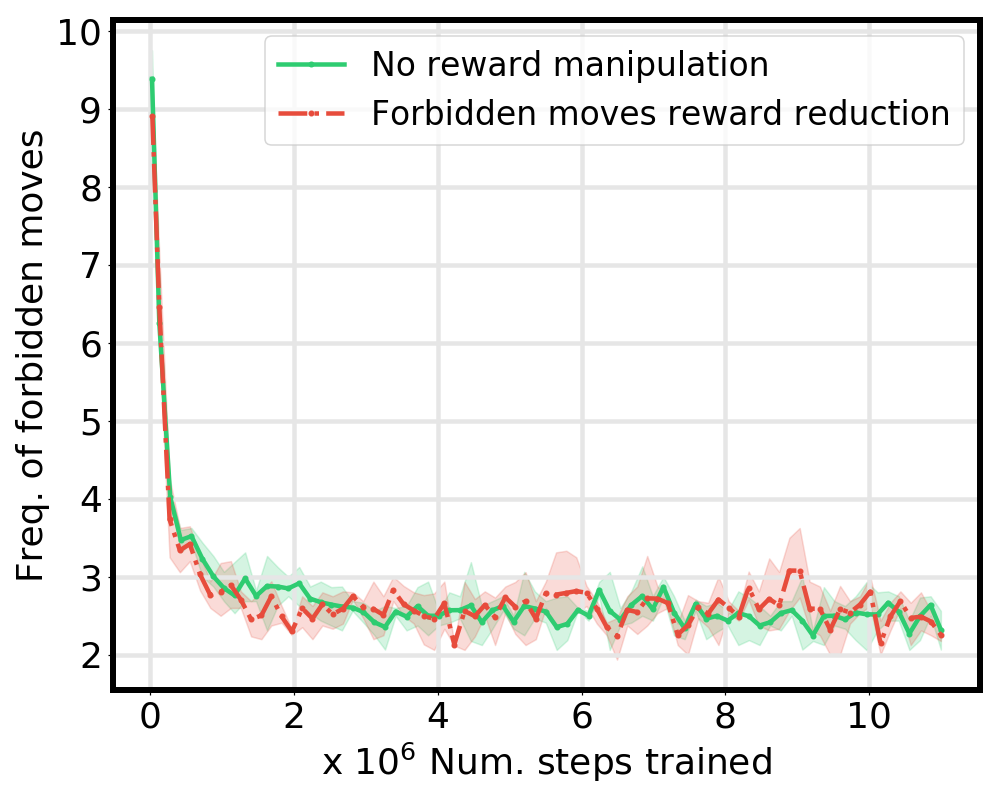}
    \caption{Experiments with reward manipulation for rule 2, forbidden moves with 5 agents on a $11 \times 11$ map.}
    \label{fig:results_RM}
\end{figure}

\textbf{Reward manipulation:}
We present the frequency of forbidden moves  (rule 2) in Figure \ref{fig:results_RM}.
At the beginning of training, the agents perform an average of 10 forbidden moves, which is the leading cause of death for the agents.
After training for around 1M time steps, the mean frequency of forbidden moves drops to around 3.
We can observe a slight reduction of the frequency of forbidden moves when comparing between the agents with and without no reward manipulation.
However, as evident from the figure, the agents also manage to learn forbidden moves avoidance even without reward manipulation.

\begin{figure}[h!]
    \centering
    \includegraphics[width=0.35\textwidth]{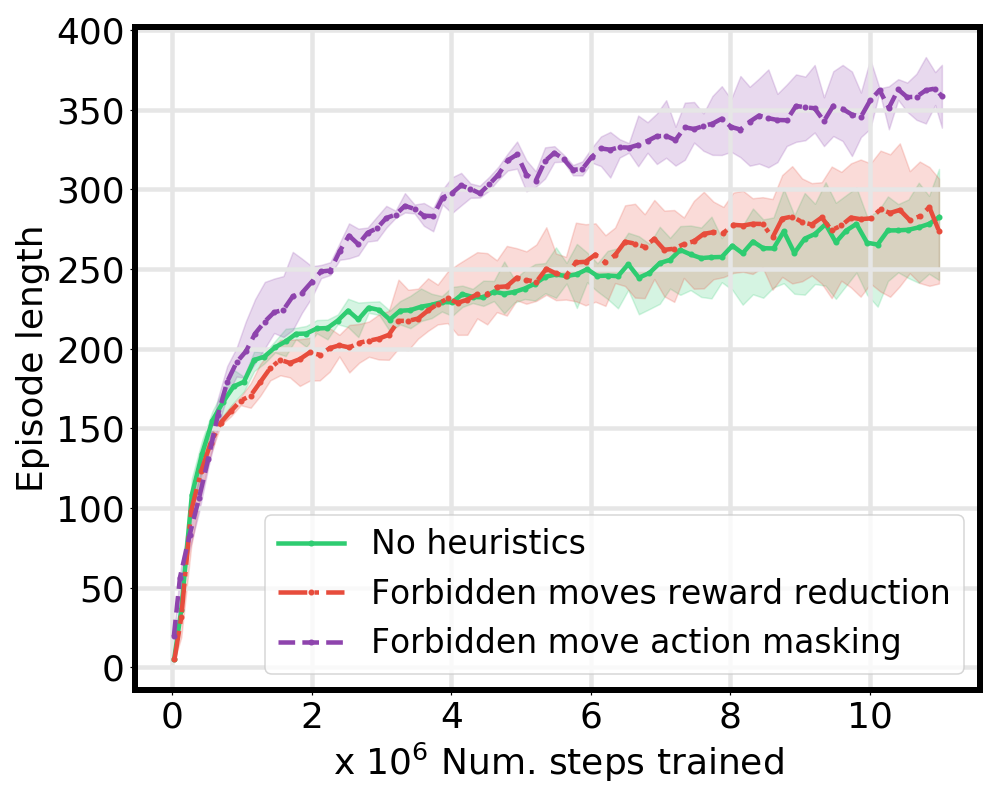}
    \caption{Comparison between the performance of in-training action masking, reward manipulation, and no HILL with 5 agents on a $11 \times 11$ map.}
    \label{fig:results_comparison}
\end{figure}

In Figure \ref{fig:results_comparison}, we can see that the episode length for agents with in-training action masking outperform agents with reward manipulation and no heuristics.
This observation is consistent with Figure \ref{fig:results_AM} where forbidden move action masking improved policy training. 

\subsection{Arena testing}

Table \ref{table:arena_score} shows the performances of the four agents in the Battlesnake arena to address rule 2. 
The four agents correspond to in-training action masking, ad-hoc action overwriting, reward manipulation, and vanilla training with no heuristics.
Each agent is trained on a $11 \times 11$ map for 2500 episodes.
The same agents are used to tested in 1 vs. 1 competition with results presented in Table \ref{table:arena_1v1}.
We observe a consistent performance from the two tables.
As expected, no HILL has the worst performance in the arena.
With mean scores of 2.533 and 2.333 for in-training action masking and ad-hoc action overwriting respectively, performance of the former is slightly higher.
This aligns with the trend shown in Figure \ref{fig:results_AM}.
It is interesting to note that the agent with reward manipulation has the best performance in the arena, suggesting possible sub-optimal actions introduced during action overwriting.

\begin{table}[h!]
\centering
\begin{tabular}{l | c c} 
 \hline
HILL type & Arena score & \%Forbidden moves  \\ [0.5ex] 
 \hline\hline
No HILL & $2.200 \pm 0.846$ & 26.6\%\\
In-training AM & $2.533 \pm 1.074$ & 0\%\\
RM & $2.900 \pm 1.296$ & 13.3\%\\
Ad-hoc AO& $2.333 \pm 1.154$ & 0 \% \\
\end{tabular}
\caption{Scores in the Battlesnake arena and the \% of deaths caused by forbidden moves. AM refers to action masking, RM refers to reward manipulation, and AO refers to action overwriting.}
\label{table:arena_score}
\end{table}

\begin{table}[h!]
\centering
\begin{tabular}{r | c c c c} 
&  No HILL & IT AM & RM & AH AO \\ [0.5ex] 
\hline
No HILL & - & 4 & 3 & 6\\
IT AM & 6 & - & 1 & 2   \\
RM & 7 & 9 & - & 1 \\
AH AO& 4 & 8 & 9 & - \\
\end{tabular}
\caption{Scores (with respect to the rows) in the Battlesnake arena in a 1 vs 1 format. 
IT AM, AH AO and RM refer to in-training action masking, Ad-hoc action overwriting and reward manipulation respectively.}
\label{table:arena_1v1}
\end{table}
\section{Conclusion}

We introduced the Battlesnake Challenge, a framework to effectively experiment and evaluate multi-agent reinforcement learning with human-in-the-loop.
We formulated Battlesnake into a multi-agent RL problem and identified a set of heuristics-based rules to facilitate standardised human-in-the-loop RL research.
We presented the performance of three heuristics incorporation methods. 
Our results suggested that the effectiveness of these methods are different during offline training and online inference.
Overall, agents with HILL perform better than agents without HILL.
During training, action masking improves the agents' survivability, leading to increased episode lengths. 
In contrast, in the Battlesnake arena, the agent with reward manipulation outperform the agent with in-training action masking.
All of the code is readily available at our git repository. 
We look forward to contributions from the researchers and developers to build new RL based Battlesnakes.


\bibliography{bib}

\begin{thebibliography}{43}
\providecommand{\natexlab}[1]{#1}
\providecommand{\url}[1]{\texttt{#1}}
\expandafter\ifx\csname urlstyle\endcsname\relax
  \providecommand{\doi}[1]{doi: #1}\else
  \providecommand{\doi}{doi: \begingroup \urlstyle{rm}\Url}\fi

\bibitem[Abel et~al.(2017)Abel, Salvatier, Stuhlm{\"u}ller, and
  Evans]{abel2017agent}
Abel, D., Salvatier, J., Stuhlm{\"u}ller, A., and Evans, O.
\newblock Agent-agnostic human-in-the-loop reinforcement learning.
\newblock \emph{arXiv preprint arXiv:1701.04079}, 2017.

\bibitem[Arakawa et~al.(2018)Arakawa, Kobayashi, Unno, Tsuboi, and
  Maeda]{arakawa2018dqn}
Arakawa, R., Kobayashi, S., Unno, Y., Tsuboi, Y., and Maeda, S.-i.
\newblock Dqn-tamer: Human-in-the-loop reinforcement learning with intractable
  feedback.
\newblock \emph{arXiv preprint arXiv:1810.11748}, 2018.

\bibitem[Arumugam et~al.(2019)Arumugam, Lee, Saskin, and
  Littman]{arumugam2019deep}
Arumugam, D., Lee, J.~K., Saskin, S., and Littman, M.~L.
\newblock Deep reinforcement learning from policy-dependent human feedback.
\newblock \emph{arXiv preprint arXiv:1902.04257}, 2019.

\bibitem[Brockman et~al.(2016)Brockman, Cheung, Pettersson, Schneider,
  Schulman, Tang, and Zaremba]{brockman2016openai}
Brockman, G., Cheung, V., Pettersson, L., Schneider, J., Schulman, J., Tang,
  J., and Zaremba, W.
\newblock Openai gym.
\newblock \emph{arXiv preprint arXiv:1606.01540}, 2016.

\bibitem[Bu et~al.(2008)Bu, Babu, De~Schutter, et~al.]{bu2008comprehensive}
Bu, L., Babu, R., De~Schutter, B., et~al.
\newblock A comprehensive survey of multiagent reinforcement learning.
\newblock \emph{IEEE Transactions on Systems, Man, and Cybernetics, Part C
  (Applications and Reviews)}, 38\penalty0 (2):\penalty0 156--172, 2008.

\bibitem[Bu{\c{s}}oniu et~al.(2010)Bu{\c{s}}oniu, Babu{\v{s}}ka, and
  De~Schutter]{bucsoniu2010multi}
Bu{\c{s}}oniu, L., Babu{\v{s}}ka, R., and De~Schutter, B.
\newblock Multi-agent reinforcement learning: An overview.
\newblock In \emph{Innovations in multi-agent systems and applications-1}, pp.\
   183--221. Springer, 2010.

\bibitem[Christiano et~al.(2017)Christiano, Leike, Brown, Martic, Legg, and
  Amodei]{christiano2017deep}
Christiano, P.~F., Leike, J., Brown, T., Martic, M., Legg, S., and Amodei, D.
\newblock Deep reinforcement learning from human preferences.
\newblock In \emph{Advances in Neural Information Processing Systems}, pp.\
  4299--4307, 2017.

\bibitem[Gao et~al.(2019)Gao, Hernandez-Leal, Kartal, and
  Taylor]{gao2019skynet}
Gao, C., Hernandez-Leal, P., Kartal, B., and Taylor, M.~E.
\newblock Skynet: A top deep rl agent in the inaugural pommerman team
  competition.
\newblock \emph{arXiv preprint arXiv:1905.01360}, 2019.

\bibitem[Griffith et~al.(2013)Griffith, Subramanian, Scholz, Isbell, and
  Thomaz]{griffith2013policy}
Griffith, S., Subramanian, K., Scholz, J., Isbell, C.~L., and Thomaz, A.~L.
\newblock Policy shaping: Integrating human feedback with reinforcement
  learning.
\newblock In \emph{Advances in neural information processing systems}, pp.\
  2625--2633, 2013.

\bibitem[Haarnoja et~al.(2018)Haarnoja, Zhou, Abbeel, and
  Levine]{haarnoja2018soft}
Haarnoja, T., Zhou, A., Abbeel, P., and Levine, S.
\newblock Soft actor-critic: Off-policy maximum entropy deep reinforcement
  learning with a stochastic actor.
\newblock \emph{arXiv preprint arXiv:1801.01290}, 2018.

\bibitem[Hausknecht et~al.(2016)Hausknecht, Mupparaju, Subramanian,
  Kalyanakrishnan, and Stone]{hausknecht2016half}
Hausknecht, M., Mupparaju, P., Subramanian, S., Kalyanakrishnan, S., and Stone,
  P.
\newblock Half field offense: An environment for multiagent learning and ad hoc
  teamwork.
\newblock In \emph{AAMAS Adaptive Learning Agents (ALA) Workshop}. sn, 2016.

\bibitem[Kalyanakrishnan et~al.(2006)Kalyanakrishnan, Liu, and
  Stone]{kalyanakrishnan2006half}
Kalyanakrishnan, S., Liu, Y., and Stone, P.
\newblock Half field offense in robocup soccer: A multiagent reinforcement
  learning case study.
\newblock In \emph{Robot Soccer World Cup}, pp.\  72--85. Springer, 2006.

\bibitem[Knox \& Stone(2008)Knox and Stone]{knox2008tamer}
Knox, W.~B. and Stone, P.
\newblock Tamer: Training an agent manually via evaluative reinforcement.
\newblock In \emph{2008 7th IEEE International Conference on Development and
  Learning}, pp.\  292--297. IEEE, 2008.

\bibitem[Knox \& Stone(2009)Knox and Stone]{knox2009interactively}
Knox, W.~B. and Stone, P.
\newblock Interactively shaping agents via human reinforcement: The tamer
  framework.
\newblock In \emph{Proceedings of the fifth international conference on
  Knowledge capture}, pp.\  9--16, 2009.

\bibitem[Knox \& Stone(2010)Knox and Stone]{knox2010combining}
Knox, W.~B. and Stone, P.
\newblock Combining manual feedback with subsequent mdp reward signals for
  reinforcement learning.
\newblock In \emph{Proceedings of the 9th International Conference on
  Autonomous Agents and Multiagent Systems: volume 1-Volume 1}, pp.\  5--12.
  Citeseer, 2010.

\bibitem[Knox \& Stone(2012)Knox and Stone]{knox2012reinforcement}
Knox, W.~B. and Stone, P.
\newblock Reinforcement learning from simultaneous human and mdp reward.
\newblock In \emph{AAMAS}, pp.\  475--482, 2012.

\bibitem[Liang et~al.(2017)Liang, Liaw, Moritz, Nishihara, Fox, Goldberg,
  Gonzalez, Jordan, and Stoica]{liang2017rllib}
Liang, E., Liaw, R., Moritz, P., Nishihara, R., Fox, R., Goldberg, K.,
  Gonzalez, J.~E., Jordan, M.~I., and Stoica, I.
\newblock Rllib: Abstractions for distributed reinforcement learning.
\newblock \emph{arXiv preprint arXiv:1712.09381}, 2017.

\bibitem[Littman(1994)]{littman1994markov}
Littman, M.~L.
\newblock Markov games as a framework for multi-agent reinforcement learning.
\newblock In \emph{Machine learning proceedings 1994}, pp.\  157--163.
  Elsevier, 1994.

\bibitem[Lowe et~al.(2017)Lowe, Wu, Tamar, Harb, Abbeel, and
  Mordatch]{lowe2017multi}
Lowe, R., Wu, Y.~I., Tamar, A., Harb, J., Abbeel, O.~P., and Mordatch, I.
\newblock Multi-agent actor-critic for mixed cooperative-competitive
  environments.
\newblock In \emph{Advances in neural information processing systems}, pp.\
  6379--6390, 2017.

\bibitem[MacGlashan et~al.(2017)MacGlashan, Ho, Loftin, Peng, Wang, Roberts,
  Taylor, and Littman]{macglashan2017interactive}
MacGlashan, J., Ho, M.~K., Loftin, R., Peng, B., Wang, G., Roberts, D.~L.,
  Taylor, M.~E., and Littman, M.~L.
\newblock Interactive learning from policy-dependent human feedback.
\newblock In \emph{Proceedings of the 34th International Conference on Machine
  Learning-Volume 70}, pp.\  2285--2294. JMLR. org, 2017.

\bibitem[Matiisen et~al.(2019)Matiisen, Oliver, Cohen, and
  Schulman]{matiisen2019teacher}
Matiisen, T., Oliver, A., Cohen, T., and Schulman, J.
\newblock Teacher-student curriculum learning.
\newblock \emph{IEEE transactions on neural networks and learning systems},
  2019.

\bibitem[Meisheri et~al.(2019)Meisheri, Shelke, Verma, and
  Khadilkar]{meisheri2019accelerating}
Meisheri, H., Shelke, O., Verma, R., and Khadilkar, H.
\newblock Accelerating training in pommerman with imitation and reinforcement
  learning.
\newblock \emph{arXiv preprint arXiv:1911.04947}, 2019.

\bibitem[Mnih et~al.(2015)Mnih, Kavukcuoglu, Silver, Rusu, Veness, Bellemare,
  Graves, Riedmiller, Fidjeland, Ostrovski, et~al.]{mnih2015human}
Mnih, V., Kavukcuoglu, K., Silver, D., Rusu, A.~A., Veness, J., Bellemare,
  M.~G., Graves, A., Riedmiller, M., Fidjeland, A.~K., Ostrovski, G., et~al.
\newblock Human-level control through deep reinforcement learning.
\newblock \emph{Nature}, 518\penalty0 (7540):\penalty0 529--533, 2015.

\bibitem[Nguyen et~al.(2020)Nguyen, Nguyen, and Nahavandi]{nguyen2020deep}
Nguyen, T.~T., Nguyen, N.~D., and Nahavandi, S.
\newblock Deep reinforcement learning for multiagent systems: A review of
  challenges, solutions, and applications.
\newblock \emph{IEEE Transactions on Cybernetics}, 2020.

\bibitem[Portelas et~al.(2019)Portelas, Colas, Hofmann, and
  Oudeyer]{portelas2019teacher}
Portelas, R., Colas, C., Hofmann, K., and Oudeyer, P.-Y.
\newblock Teacher algorithms for curriculum learning of deep rl in continuously
  parameterized environments.
\newblock \emph{arXiv preprint arXiv:1910.07224}, 2019.

\bibitem[Rashid et~al.(2018)Rashid, Samvelyan, De~Witt, Farquhar, Foerster, and
  Whiteson]{rashid2018qmix}
Rashid, T., Samvelyan, M., De~Witt, C.~S., Farquhar, G., Foerster, J., and
  Whiteson, S.
\newblock Qmix: monotonic value function factorisation for deep multi-agent
  reinforcement learning.
\newblock \emph{arXiv preprint arXiv:1803.11485}, 2018.

\bibitem[Resnick et~al.(2018)Resnick, Eldridge, Ha, Britz, Foerster, Togelius,
  Cho, and Bruna]{resnick2018pommerman}
Resnick, C., Eldridge, W., Ha, D., Britz, D., Foerster, J., Togelius, J., Cho,
  K., and Bruna, J.
\newblock Pommerman: A multi-agent playground.
\newblock \emph{arXiv preprint arXiv:1809.07124}, 2018.

\bibitem[Russell \& Norvig(2002)Russell and Norvig]{russell2002artificial}
Russell, S. and Norvig, P.
\newblock Artificial intelligence: a modern approach.
\newblock 2002.

\bibitem[Samvelyan et~al.(2019)Samvelyan, Rashid, Schroeder~de Witt, Farquhar,
  Nardelli, Rudner, Hung, Torr, Foerster, and Whiteson]{samvelyan2019starcraft}
Samvelyan, M., Rashid, T., Schroeder~de Witt, C., Farquhar, G., Nardelli, N.,
  Rudner, T.~G., Hung, C.-M., Torr, P.~H., Foerster, J., and Whiteson, S.
\newblock The starcraft multi-agent challenge.
\newblock In \emph{Proceedings of the 18th International Conference on
  Autonomous Agents and MultiAgent Systems}, pp.\  2186--2188. International
  Foundation for Autonomous Agents and Multiagent Systems, 2019.

\bibitem[Saunders et~al.(2018)Saunders, Sastry, Stuhlmueller, and
  Evans]{saunders2018trial}
Saunders, W., Sastry, G., Stuhlmueller, A., and Evans, O.
\newblock Trial without error: Towards safe reinforcement learning via human
  intervention.
\newblock In \emph{Proceedings of the 17th International Conference on
  Autonomous Agents and MultiAgent Systems}, pp.\  2067--2069. International
  Foundation for Autonomous Agents and Multiagent Systems, 2018.

\bibitem[Schier \& Wüstenbecker(2019)Schier and Wüstenbecker]{mci/Schier2019}
Schier, M.~B. and Wüstenbecker, N.
\newblock Adversarial n-player search using locality for the game of
  battlesnake.
\newblock In Becker, M. (ed.), \emph{SKILL 2019 - Studierendenkonferenz
  Informatik}, pp.\  109--120, Bonn, 2019. Gesellschaft für Informatik e.V.

\bibitem[Schulman et~al.(2015)Schulman, Moritz, Levine, Jordan, and
  Abbeel]{schulman2015high}
Schulman, J., Moritz, P., Levine, S., Jordan, M., and Abbeel, P.
\newblock High-dimensional continuous control using generalized advantage
  estimation.
\newblock \emph{arXiv preprint arXiv:1506.02438}, 2015.

\bibitem[Schulman et~al.(2017)Schulman, Wolski, Dhariwal, Radford, and
  Klimov]{schulman2017proximal}
Schulman, J., Wolski, F., Dhariwal, P., Radford, A., and Klimov, O.
\newblock Proximal policy optimization algorithms.
\newblock \emph{arXiv preprint arXiv:1707.06347}, 2017.

\bibitem[Silver et~al.(2017)Silver, Schrittwieser, Simonyan, Antonoglou, Huang,
  Guez, Hubert, Baker, Lai, Bolton, et~al.]{silver2017mastering}
Silver, D., Schrittwieser, J., Simonyan, K., Antonoglou, I., Huang, A., Guez,
  A., Hubert, T., Baker, L., Lai, M., Bolton, A., et~al.
\newblock Mastering the game of go without human knowledge.
\newblock \emph{Nature}, 550\penalty0 (7676):\penalty0 354--359, 2017.

\bibitem[Stone et~al.(2005)Stone, Kuhlmann, Taylor, and Liu]{stone2005keepaway}
Stone, P., Kuhlmann, G., Taylor, M.~E., and Liu, Y.
\newblock Keepaway soccer: From machine learning testbed to benchmark.
\newblock In \emph{Robot Soccer World Cup}, pp.\  93--105. Springer, 2005.

\bibitem[Vinyals et~al.(2019)Vinyals, Babuschkin, Czarnecki, Mathieu, Dudzik,
  Chung, Choi, Powell, Ewalds, Georgiev, et~al.]{vinyals2019grandmaster}
Vinyals, O., Babuschkin, I., Czarnecki, W.~M., Mathieu, M., Dudzik, A., Chung,
  J., Choi, D.~H., Powell, R., Ewalds, T., Georgiev, P., et~al.
\newblock Grandmaster level in starcraft ii using multi-agent reinforcement
  learning.
\newblock \emph{Nature}, 575\penalty0 (7782):\penalty0 350--354, 2019.

\bibitem[Warnell et~al.(2018)Warnell, Waytowich, Lawhern, and
  Stone]{warnell2018deep}
Warnell, G., Waytowich, N., Lawhern, V., and Stone, P.
\newblock Deep tamer: Interactive agent shaping in high-dimensional state
  spaces.
\newblock In \emph{Thirty-Second AAAI Conference on Artificial Intelligence},
  2018.

\bibitem[Xiao et~al.(2020)Xiao, Lu, Ramasubramanian, Clark, Bushnell, and
  Poovendran]{xiao2020fresh}
Xiao, B., Lu, Q., Ramasubramanian, B., Clark, A., Bushnell, L., and Poovendran,
  R.
\newblock Fresh: Interactive reward shaping in high-dimensional state spaces
  using human feedback.
\newblock \emph{arXiv preprint arXiv:2001.06781}, 2020.

\bibitem[Yang et~al.(2018)Yang, Luo, Li, Zhou, Zhang, and Wang]{yang2018mean}
Yang, Y., Luo, R., Li, M., Zhou, M., Zhang, W., and Wang, J.
\newblock Mean field multi-agent reinforcement learning.
\newblock \emph{arXiv preprint arXiv:1802.05438}, 2018.

\bibitem[Ye et~al.(2019)Ye, Liu, Sun, Shi, Zhao, Wu, Yu, Yang, Wu, Guo,
  et~al.]{ye2019mastering}
Ye, D., Liu, Z., Sun, M., Shi, B., Zhao, P., Wu, H., Yu, H., Yang, S., Wu, X.,
  Guo, Q., et~al.
\newblock Mastering complex control in moba games with deep reinforcement
  learning.
\newblock \emph{arXiv preprint arXiv:1912.09729}, 2019.

\bibitem[Zhang et~al.(2019{\natexlab{a}})Zhang, Yang, and
  Ba{\c{s}}ar]{zhang2019multi}
Zhang, K., Yang, Z., and Ba{\c{s}}ar, T.
\newblock Multi-agent reinforcement learning: A selective overview of theories
  and algorithms.
\newblock \emph{arXiv preprint arXiv:1911.10635}, 2019{\natexlab{a}}.

\bibitem[Zhang et~al.(2019{\natexlab{b}})Zhang, Torabi, Guan, Ballard, and
  Stone]{zhang2019leveraging}
Zhang, R., Torabi, F., Guan, L., Ballard, D.~H., and Stone, P.
\newblock Leveraging human guidance for deep reinforcement learning tasks.
\newblock \emph{arXiv preprint arXiv:1909.09906}, 2019{\natexlab{b}}.

\bibitem[Zheng et~al.(2018)Zheng, Yang, Cai, Zhou, Zhang, Wang, and
  Yu]{zheng2018magent}
Zheng, L., Yang, J., Cai, H., Zhou, M., Zhang, W., Wang, J., and Yu, Y.
\newblock Magent: A many-agent reinforcement learning platform for artificial
  collective intelligence.
\newblock In \emph{Thirty-Second AAAI Conference on Artificial Intelligence},
  2018.

\end{thebibliography}
\bibliographystyle{icml2020}


%



\end{document}